\documentclass{article}
\usepackage{ijcai26}

\usepackage{times}
\usepackage{soul}
\usepackage{url}
\usepackage[hidelinks]{hyperref}
\usepackage[utf8]{inputenc}
\usepackage[small]{caption}
\usepackage{graphicx}
\usepackage{amsmath}
\usepackage{amsthm}
\usepackage{booktabs}
\usepackage{algorithm}
\usepackage{algorithmic}
\usepackage[switch]{lineno}
\usepackage{multirow}

\usepackage{booktabs}
\usepackage[table]{xcolor}
\usepackage{adjustbox}

\usepackage{amsfonts}

\usepackage{xcolor}

\newcommand{\gray}[1]{\textcolor{gray}{#1}}

\definecolor{blockpurple}{RGB}{240,235,255}
\definecolor{blockpink}{RGB}{255,235,240}
\definecolor{blockyellow}{RGB}{255,248,230}
\definecolor{blockblue}{RGB}{225,240,255}

\title{GRPO-TTA: Test-Time Visual Tuning for Vision-Language Models via GRPO-Driven Reinforcement Learning}

\author{
    Yujun Li$^1$ \and
    Hongyuan Zhang$^{2,3}$ \and
    Yuan Yuan$^1$\\
    \affiliations
    $^1$ School of Artificial Intelligence, Optics and Electronics (iOPEN),\\
    Northwestern Polytechnical University, Xi'an 710072, P.R. China\\
    $^2$ The University of Hong Kong\\
    \emails{
        yujunli361@gmail.com,
        hyzhang98@gmail.com,
        y.yuan1.ieee@gmail.com (corresponding author)
    }
}

\begin{document}

\maketitle

\begin{abstract}
    Group Relative Policy Optimization (GRPO) has recently shown strong performance in post-training large language models and vision-language models. It raises a question of whether the GRPO also significantly promotes the test-time adaptation (TTA) of vision–language models. In this paper, we propose Group Relative Policy Optimization for Test-Time Adaptation (GRPO-TTA), which adapts GRPO to the TTA setting by reformulating class-specific prompt prediction as a group-wise policy optimization problem. Specifically, we construct output groups by sampling top-$K$ class candidates from CLIP similarity distributions, enabling probability-driven optimization without access to ground-truth labels. Moreover, we design reward functions tailored to test-time adaptation, including alignment rewards and dispersion rewards, to guide effective visual encoder tuning. Extensive experiments across 15 diverse benchmarks demonstrate that GRPO-TTA consistently outperforms existing test-time adaptation methods, with notably larger performance gains under natural distribution shifts.
\end{abstract}

\section{Introduction}

In recent years, large-scale vision–language pre-trained models (VLMs) have driven remarkable progress in computer vision and multimodal learning ~\cite{radford2021learning,zhang2025unified}. Notably, CLIP ~\cite{radford2021learning}, trained on 400 million image–text pairs, has demonstrated strong zero-shot generalization across a wide range of downstream tasks. This success demonstrates the potential of VLMs as a versatile foundation model, sparking growing interest in their transfer learning capabilities. Prior studies mainly focus on adapting VLMs through prompt tuning ~\cite{zhou2022learning} or visual modifications such as adapters ~\cite{gao2024clip} and lightweight fine-tuning modules ~\cite{huang2025enhance}. Despite effectiveness, these approaches suffer from two inherent limitations ~\cite{zhou2022conditional,lafon2025cliptta}: 
\begin{figure}[htbp]
\centering
\includegraphics[scale=0.20]{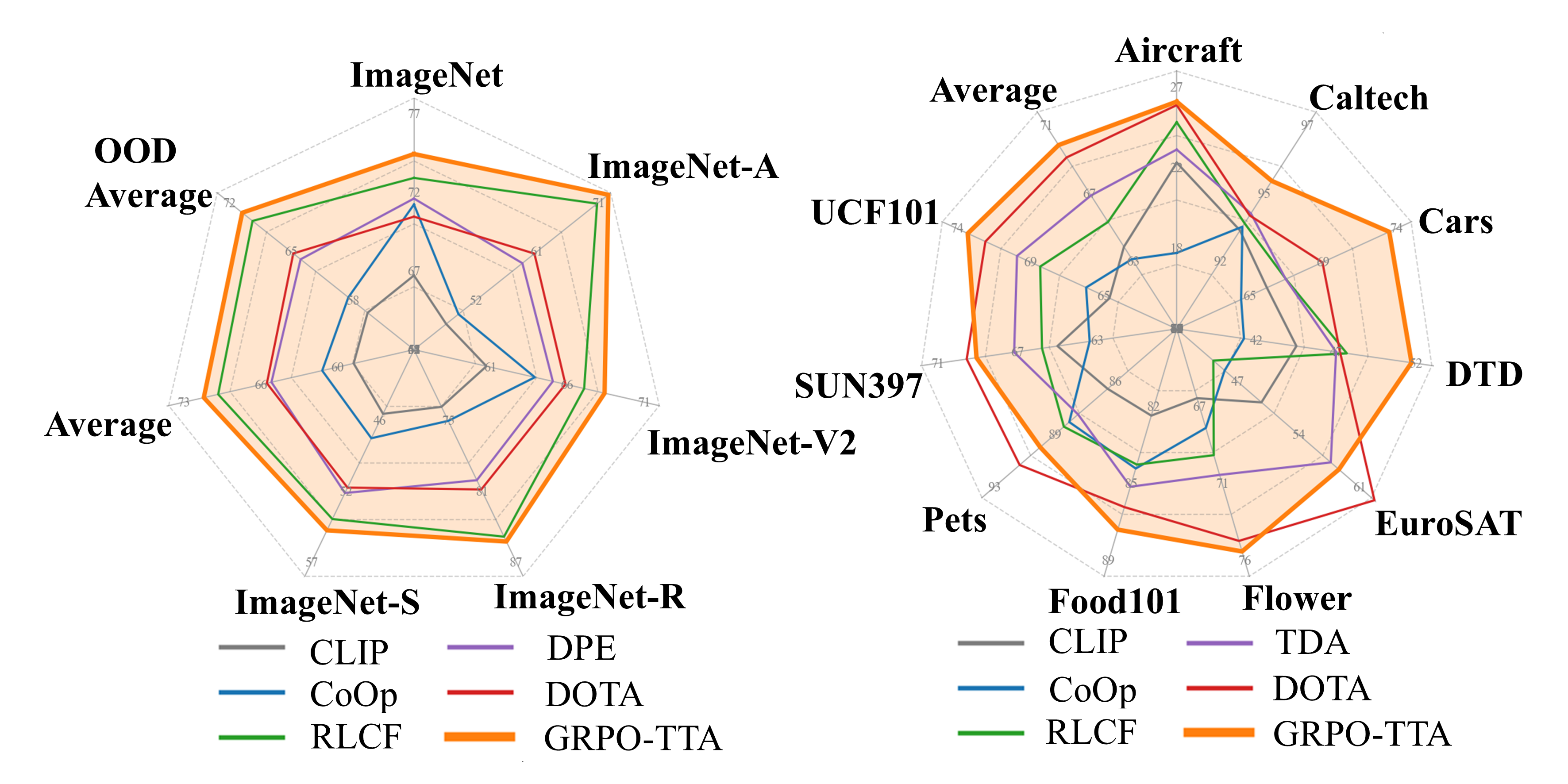}
	\caption{\textbf{Performance comparison of different methods.} Across ImageNet variants {\it (left)} and cross-domain datasets {\it (right)}, GRPO-TTA consistently surpasses existing test-time adaptation methods.}
	\label{visual1}
\end{figure}
(1) the zero-shot generalization ability of VLMs often degrades severely under distribution shifts between pre-training data and downstream test data, and (2) these approaches rely on access to task-specific training data or labeled validation sets, which is often impractical in real-world scenarios.

To address these challenges, test-time adaptation (TTA) has emerged as a promising paradigm that enables models to adapt dynamically during inference without access to labeled data ~\cite{shu2022test,feng2023diverse,zhao2024test,lafon2025cliptta}. In the context of VLMs, TTA methods typically exploit multiple augmented views of each test image and perform sample-wise optimization of either textual prompts or visual representations. By adapting the model online to the test distribution, TTA offers an effective mechanism for mitigating domain shift.

Recently, the release of DeepSeek-R1 ~\cite{guo2025deepseek} has popularized a new optimization paradigm known as Group Relative Policy Optimization (GRPO). As a reinforcement learning method, GRPO has shown strong effectiveness in fine-tuning large language models (LLMs) ~\cite{yu2025dapo} and vision–language models (VLMs) ~\cite{shen2025vlm} by optimizing relative performance within groups of sampled outputs. This naturally raises an interesting question: \textbf{\textit{can GRPO be extended beyond post-training and applied to test-time adaptation of VLMs?}} In this work, we are going to answer the above question and propose a GRPO-based framework customized for TTA.

The backbone architectures of VLMs vary across different methods, with convolutional networks such as ResNet ~\cite{he2016deep}, and Vision Transformers (ViTs) ~\cite{dosovitskiy2020image} being representative examples. Notably, both LLMs and ViT-based visual encoders are built upon transformer architectures, suggesting a natural architectural compatibility between GRPO and ViT-based VLMs. Motivated by this observation, we focus on adapting GRPO to TTA in ViT-based VLMs. Specifically, building upon the augmentation and confidence selection pipeline of TPT ~\cite{shu2022test}, we propose to tune the projector that projects the image into the LLM token embedding space. For each test sample, we construct a group of candidate outputs by sampling $K$ responses from the model’s output distribution.

To enable group-wise optimization applied to TTA, our proposed method treats the prediction distribution over class-specific textual prompts as a structured output space. Unlike conventional GRPO settings that rely on autoregressive token generation and explicit supervision, our formulation leverages the inherent probabilistic similarity scores produced by CLIP, which allows GRPO to operate without access to ground-truth labels. Moreover, we redesigned the reward function to implement GRPO generalized to TTA effectively. In particular, we propose a reward decomposition strategy that balances alignment and distributional dispersion in the similarity score space.

We extensively evaluate the test-time generalization performance of the proposed GRPO-TTA across 15 diverse recognition benchmarks in two scenarios: the cross-domain generalization scenario and the natural distribution shift scenario. As shown in Figure \ref{visual1}, GRPO-TTA consistently outperforms existing TTA methods across diverse tasks. The contributions of this paper are summarized as follows:

\begin{itemize}
\item Based on the characteristics of test-time adaptation and vision–language models, we reformulate the GRPO objective by redesigning reward functions.  We introduce an \textbf{alignment reward} and a \textbf{dispersion reward} to guide sample-wise adaptation, encouraging the model to align visual representations with semantically relevant textual candidates while avoiding collapsed predictions during TTA.
\item To the best of our knowledge, GRPO-TTA is the first work that successfully adapts GRPO to the test-time adaptation setting, enabling effective sample-level optimization without access to ground-truth labels.
\item Extensive experiments on 15 diverse benchmarks demonstrate that the proposed method consistently outperforms existing test-time adaptation approaches, achieving $1.21\%\mathord{\sim} 11.19\%$ improvement under natural distribution shifts and an average gain of $0.65\%\mathord{\sim} 5.53\%$ on cross-domain benchmarks, while introducing only a modest additional computational overhead.
\end{itemize}

\section{Related Work}

\begin{figure*}[htbp]
\centering
\includegraphics[scale=0.39]{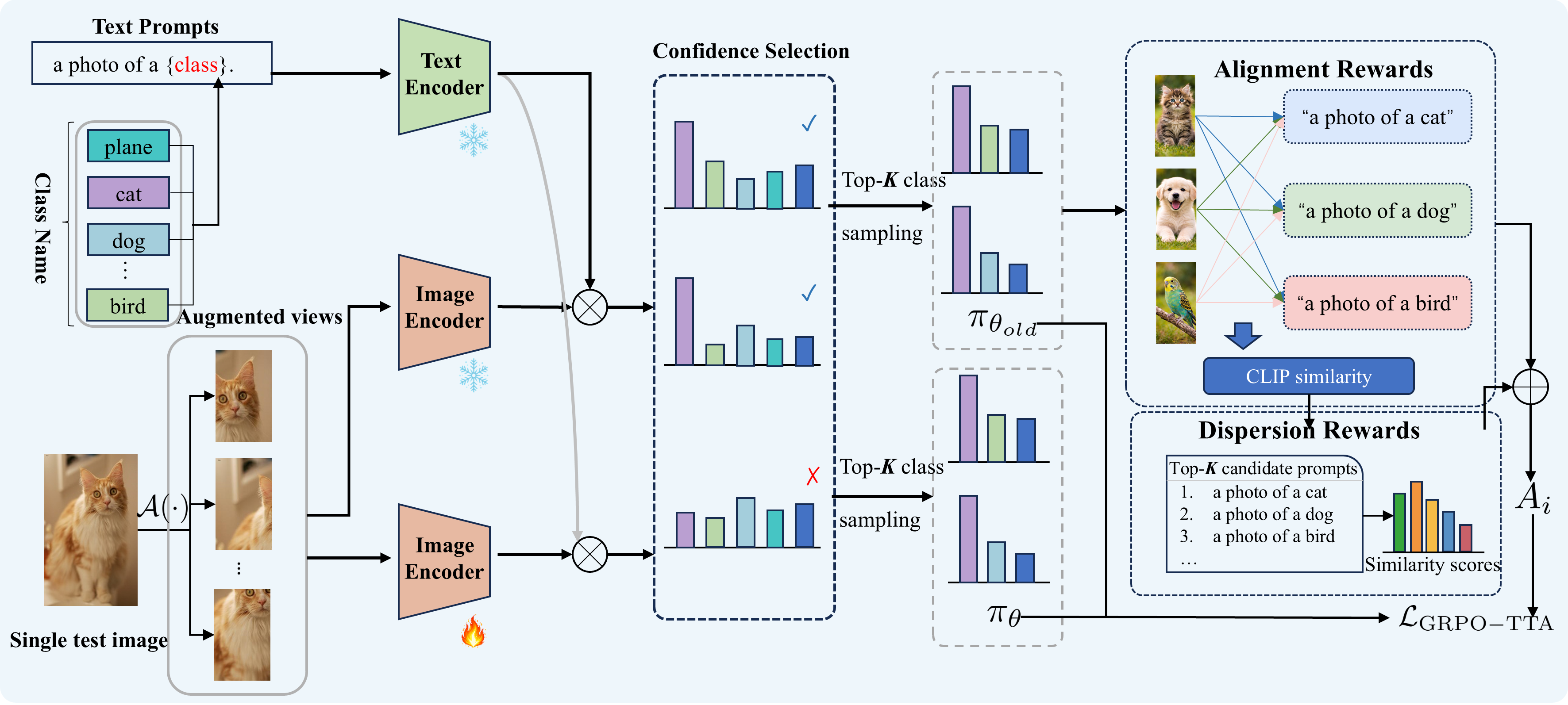}
\caption{\textbf{The framework of the proposed GRPO-TTA framework.} Firstly, a single test image is augmented to generate multiple views, where confident views with low-entropy predictions are selected. Then, for each selected view, we sample the top-$K$ classes and calculate their probabilistic distributions. Finally, advantages are computed using CLIP similarity and the probabilistic distributions generated by the similarity scores.}
\label{figure1}
\end{figure*}

\subsection{Vision-Language Models}
Vision–Language Models (VLMs) leverage large-scale image–text pre-training to learn aligned multimodal representations, among which CLIP ~\cite{radford2021learning} has demonstrated strong zero-shot generalization. To adapt pretrained VLMs to downstream tasks, existing approaches mainly focus on prompt-based methods ~\cite{zhou2022learning,zhou2022conditional} and lightweight visual adaptation modules such as adapters ~\cite{gao2024clip} or noise-based tuning ~\cite{huang2025enhance}. Although effective, most of these methods require access to labeled target data or additional training stages, which limits their applicability in practical scenarios.

\subsection{Test-Time Adaptation}
Test-time adaptation (TTA) has emerged as a crucial paradigm for addressing domain shift by enabling models to adapt dynamically using only unlabeled test samples. A widely adopted strategy minimizes prediction entropy either across a batch of test samples ~\cite{lafon2025cliptta} or across multiple augmented views of a single test image ~\cite{zhang2024dual}. TPT ~\cite{shu2022test} was the first to extend TTA to vision–language models by fine-tuning textual prompts through consistency across augmented views of each test sample. DiffTPT ~\cite{feng2023diverse} further enhanced this framework by introducing diffusion-based image generation to enrich the augmentation pool. However, TPT and DiffTPT require backpropagation through large pretrained models and repeated inference over multiple augmented views, making them computationally expensive and resource-intensive. TDA ~\cite{karmanov2024efficient} proposes a lightweight test-time adaptation method by storing representative test samples. DOTA ~\cite{han2024dota} further enhanced this framework by continuously estimating the distribution of test samples. However, these approaches either incur high computational costs or exhibit limited robustness under natural distribution shifts. In contrast, our method explores reinforcement-based test-time adaptation tailored to VLMs.

\subsection{Reinforcement Learning in Language and Vision}
Reinforcement learning (RL) has become a cornerstone technique in the post-training of large language models (LLMs), playing a pivotal role in aligning model behaviors with human preferences and task objectives ~\cite{lee2023rlaif}. The most representative paradigm is reinforcement learning with human feedback (RLHF) ~\cite{ouyang2022training}. The reward model trained from human preference annotations is used to fine-tune the policy model, typically via Proximal Policy Optimization (PPO) ~\cite{schulman2017proximal}. Beyond PPO-based frameworks, alternative optimization strategies have been proposed to reduce training complexity. Direct Preference Optimization (DPO) ~\cite{rafailov2023direct} reformulates preference learning as a classification problem, directly optimizing the policy without explicitly training or querying a reward model. More recently, Group Relative Policy Optimization (GRPO) ~\cite{guo2025deepseek} has emerged as an effective reinforcement learning method for post-training LLMs. Different from PPO and DPO, GRPO samples multiple outputs for each input and computes group-wise relative advantages to guide policy updates. By eliminating the need for generalized advantage estimation, GRPO substantially reduces computational overhead while maintaining stable optimization dynamics. Moreover, GRPO has recently been extended to broader domains beyond LLMs. VLM-R1 ~\cite{shen2025vlm} introduces GRPO into vision–language models, and GRPO-RM ~\cite{xu2025grpo} further applies it to representation learning in computer vision. In this work, we take a step further by investigating GRPO-based test-time adaptation for VLMs. For efficiency and practical deployment, we only focus on visual encoder tuning during TTA.

\section{Methodology}

In this section, we first introduce the GRPO adaptation for test-time adaptation of VLMs. Then, with a redesigned reward function, the GRPO-TTA method is elaborated. The framework of our method is illustrated in Figure \ref{figure1}.

\subsection{Preliminaries}

\noindent{\bf{Zero-Shot Classification.}} CLIP consists of an image encoder $f_{\phi_{v}}^{v}(\cdot)$ that maps the image into a feature vector, and a text encoder $f_{\phi_{t}}^{t}(\cdot)$ that maps the text input into a feature vector. Given an input image $x$ and a textual description $t$, CLIP produces its corresponding normalized representations: $H_v=f_{\phi_{v}}^{v}(x)$ and $H_t=f_{\phi_{t}}^{t}(t)$. For a downstream image classification task involving $C$ categories, CLIP adopts a prompt-based inference strategy. Specifically, a textual template is instantiated for each class, such as "a photo of a $\texttt{<class name>}$", producing a set of class-specific text prompts $\{t_1,\cdots,t_C\}$. Each visual embedding $H_v$ is computed with all class text embeddings using cosine similarity, scaled by a temperature parameter $\tau$. The resulting similarity scores are converted into a probability distribution over classes via the softmax function:
\begin{equation}
	p\left(t_c \mid x\right) = \frac{\exp \left(\cos\left(H_{v},H_{t_c}\right) / \tau \right)}
					{\sum_{j=1}^{C} \exp\left( \cos\left(H_{v},H_{t_j}\right) / \tau \right)        }.
\label{eq:clip}
\end{equation}


\noindent{\bf{Test-Time Augmentation.}} Augmentation-based test-time adaptation approaches construct multiple views from a single input sample. Given an input image $x$, a predefined augmentation operator $\mathcal{A}(\cdot)$ is applied to generate a set of transformed instances $\{x^1,x^2,\cdots,x^n\}$, where $n$ denotes the number of augmented views. The augmentation $\mathcal{A}(\cdot)$ varies across different methods, such as AugMix ~\cite{shu2022test} and diffusion-based augmentations ~\cite{feng2023diverse}.
For the augmented views $\{x^i\}_{i=1}^n$, we can compute their zero-shot probabilities $p\left(t_c \mid x^i\right)$ with Eq. \eqref{eq:clip}. Since test-time adaptation is performed without access to ground-truth labels, prediction reliability is typically assessed through uncertainty-based criteria. A commonly adopted measure is Shannon entropy: $\mathcal{H}(x^i)=-\sum_{c=1}^{C}p\left(t_c \mid x^i\right)\log p\left(t_c \mid x^i\right)$ to quantify the confidence of the prediction. Augmented views exhibiting lower entropy are regarded as more reliable and are therefore retained for subsequent processing, while high-entropy predictions are discarded, as:
\begin{equation}
\resizebox{.91\linewidth}{!}{$
\displaystyle
	p\left(t_c \mid x^i\right) = \frac{\exp \left(\cos\left(H_{v}^i,H_{t_c}\right) / \tau \right)}
					{\sum_{j=1}^{C} \exp\left( \cos\left(H_{v}^i,H_{t_j}\right) / \tau \right)        }, \mathcal{H}(x^i)< \rho,
        $}
\end{equation}
where $\rho$ is the hyperparameter that controls the number of augmented views with lower entropy.

\noindent{\bf{Group Relative Policy Optimization.}} Given a question $q$, GRPO samples a group of outputs $\{o_1,o_2,\cdots,o_G\}$ from the old policy $\pi_{\theta_{old}}$. The policy $\pi_{\theta}$ is updated by maximizing a group-wise surrogate objective defined over these samples, where the hyper-parameters $\epsilon$ and $\beta$ are introduced to ensure stable optimization. The objective of the GRPO is as follows:
\begin{align}
 & \mathcal{J}_{G R P O}(\theta)= \mathbb{E}[q \sim P(Q),\left\{o_i\right\}_{i=1}^G \sim \pi_{\theta_{\text {old }}}(O \mid q)]  \nonumber \\
 &  \frac{1}{G}\sum_{i=1}^{G}(\tilde{A}(\theta)) \left.-\beta \mathbb{D}_{K L}\left(\pi_\theta \| \pi_{\text {ref }}\right)\right),
\label{eq:grpo}
\end{align}
with
\begin{equation}
\tilde{A}_i(\theta)=\min \left\{\frac{\pi_\theta\left(o_i \mid q\right)}{\pi_{\theta_{\mathrm{old}}}\left(o_i \mid q\right)} A_i, \operatorname{cap}_\epsilon\left(\frac{\pi_\theta\left(o_i \mid q\right)}{\pi_{\theta_{\mathrm{old}}}\left(o_i \mid q\right)}\right) A_i\right\},
\label{eq:grpo_adv}
\end{equation}
where $A_{i}$ denotes the advantage, and the fuction $\operatorname{cap}_\epsilon(z)$ outputs $z$ if $1-\epsilon \leq z \leq 1+\epsilon, 1-\epsilon$ if $z<1-\epsilon$, and $1+\epsilon$ if $z > 1+\epsilon$.

\subsection{Outputs Sampling for Single Input}
As mentioned in Section 3.1, the objective of GRPO relies on an output group with a well-defined probabilistic structure. In particular, the construction of the output group $\{o_1,o_2,\cdots,o_G\}$ directly affects the computation of the advantages $A_i$, which are central to stable group-wise optimization. To satisfy these requirements, the output sampling mechanism enables the model to generate multiple potential outputs from a single input. Each output must be linked to a specific probability and facilitate reward evaluation.
In LLMs, this requirement is naturally met through token-level sequence generation, where responses can be sampled easily. Similarly, recent VLMs ~\cite{shen2025vlm} adopt structured multimodal reasoning tasks that produce various outputs with associated probabilities. However, the test-time adaptation (TTA) setting for vision–language models introduces difficulties.

During test-time adaptation, ground-truth labels are unavailable. This makes it difficult to formulate the complex and multimodal reasoning objectives, which are commonly used in previous GRPO-based VLMs. To address this limitation, we propose a simplified yet effective alternative. Specifically, we treat a test image as a question about its category, and the complete set of class-specific textual prompts constitutes the space of all possible responses. Under this formulation, the similarity-based prediction distribution produced by the vision–language model naturally serves as the policy $\pi_{\theta}$.

The well-defined response space enables reliable estimation of relative advantages for GRPO-based optimization. Thus, to further constrain the output space and construct an output group suitable for group-wise optimization, we select the top-$K$ candidates based on the output probability distribution. Note that our method employs an additional Softmax layer generate probabilistic distributions over $K$ candidates.

\subsection{Why Using GRPO for TTA?}
The advantages formulation in GRPO is not directly applicable to test-time adaptation (TTA). In standard GRPO, the advantage $A_i$ in Eq. \eqref{eq:grpo_adv} is computed from the rewards $\{r_1,r_2,\cdots,r_G\}$ associated with a group of sampled outputs:
\begin{equation}
	A_i = \frac{r_i - \mathrm{mean}(\{r_1,r_2,\cdots,r_G\})}
		{ \mathrm{std}(\{r_1,r_2,\cdots,r_G\})}.
\label{a_1}
\end{equation}
This formulation assumes that rewards are explicitly defined for each output. This is usually achieved in post-training settings for LLMs and VLMs by setting up task-specific objectives or providing supervised feedback. In contrast, test-time adaptation operates without access to ground-truth labels. To adapt the rewards to TTA, the functions must be redesigned to accommodate their properties. In our proposed model, reward functions should satisfy two principles:
\begin{itemize}
\item text predictions corresponding to each input image cluster closely in the embedding space;
\item the model should maintain sufficient diversity among candidate outputs, avoiding collapsed predictions.
\end{itemize}

\begin{algorithm}[tb]
    \caption{Pseudo code of GRPO-TTA}
    \label{alg:algorithm}
    \textbf{Input}: Dataset $\mathcal{D}$, pretrained model $f_{pre}$\\
    \textbf{Parameter}: $K$, $\beta = 0$, $\epsilon$\\
    \textbf{Output}: $\pi_{\theta}$
    \begin{algorithmic}[1] 
        \STATE build the post-train model $\pi_{\theta}$ with $f_{pre}$ via top-$K$ classes sampling.
        \FOR {each test sample $x_i$}
        \STATE $\pi_{\theta_{o l d}} \leftarrow \pi_\theta \gray{ / /}$ \gray{copy old model}
        \STATE Generate probabilistic distribution: $\pi_{\theta_{old}}(t \mid x)$
	   \STATE Compute Rewards according to Eq. \eqref{r_1}, Eq. \eqref{r_2} and Eq. \eqref{r_3}.
	   \STATE Compute advantages according to Eq. \eqref{a_1}.
	   \STATE Compute loss $\mathcal{L}$ according to Eq. \eqref{loss} with advantages, hyper-parameters, $\pi_{\theta}$, and $\pi_{\theta_{old}}$.
	   \STATE Update parameters to minimize $\mathcal{L}$.
        \ENDFOR
        \STATE \textbf{return} $\pi_{\theta}$
    \end{algorithmic}
\end{algorithm}

\begin{table*}[]
\centering
\renewcommand{\arraystretch}{1.0}
\begin{adjustbox}{width=0.98\textwidth}
\begin{tabular}{lccccccccccc}
\toprule
Method     & Aircraft & Caltech & Cars  & DTD   & EuroSAT & Flower & Food101 & Pets  & SUN397 & UCF101 & Average        \\ \midrule
CLIP$^{\dagger}$ & 23.22    & 93.55   & 66.11 & 45.04 & 50.42   & 66.99  & 82.86   & 86.92 & 65.63  & 65.16  & 64.59          \\
CoOp$^{\dagger}$      & 18.47    & 93.70   & 64.51 & 41.92 & 46.39   & 68.71  & 85.30   & 89.14 & 64.15  & 66.55  & 63.88          \\
CoCoOp$^{\dagger}$    & 22.29    & 93.79   & 64.90 & 45.45 & 39.23   & 70.85  & 83.97   & 90.46 & 66.89  & 68.44  & 64.63          \\ \midrule
TPT$^{\dagger}$       & 24.78    & 94.16   & 66.87 & 47.75 & 42.44   & 68.98  & 84.67   & 87.79 & 65.50  & 68.04  & 65.10          \\
DiffTPT$^{\dagger}$   & 25.60    & 92.49   & 67.01 & 47.00 & 43.13   & 70.10  & \ul{87.23}   & 88.22 & 65.74  & 62.67  & 64.92          \\
RLCF$^{\ddagger}$      & 25.35    & 93.84   & 67.29 & 48.04 & 45.15   & 70.28  & 85.12   & 89.45 & 66.34  & 69.28  & 66.01          \\
WATT$^{\dagger}$      & 24.24    & 93.59   & 66.78 & 46.87 & 51.95   & 68.53  & 84.81   & 88.14 & 65.84  & 65.74  & 65.65          \\
CLIPTTA$^{\dagger}$   & \ul{26.50}    & 94.20   & 66.70 & 46.50 & \textbf{80.30}   & 71.30  & 86.70   & \ul{91.60} & 65.20  & 69.30  & \ul{69.80}          \\ \midrule
TDA$^{\dagger}$       & 23.91    & 94.24   & 67.28 & 47.40 & 58.00   & 71.42  & 86.14   & 88.63 & 67.62  & 70.66  & 67.53          \\
DPE$^{\dagger}$       & \textbf{28.95}    & \ul{94.81}   & 67.31 & \textbf{54.20} & 55.79   & 75.07  & 86.17   & 91.14 & \textbf{70.07}  & 70.44  & 69.40          \\
DOTA$^{\dagger}$      & 26.25    & 94.16   & \ul{69.56} & 47.64 & \ul{62.78}   & \ul{75.23}  & 87.08   & \textbf{92.01} & \ul{69.80}  & \ul{72.54}  & 69.71          \\ \midrule
\rowcolor{gray!15}
GRPO-TTA  & 26.43    & \textbf{95.56}   & \textbf{74.05} & \ul{51.89} & 58.89   & \textbf{75.84}  & \textbf{88.12}   & 90.82 & 69.35  & \textbf{73.59}  & \textbf{70.45} \\ \bottomrule
\end{tabular}
\end{adjustbox}
\caption{Comparison of GRPO-TTA in cross-domain generalization using ViT-B/16 as the backbone. Best results are highlighted in \textbf{bold}, and second-best results are \ul{underlined}. Results annotated with $^{\dagger}$ correspond to results reported in the original papers, whereas $^{\ddagger}$ denote results obtained using our optimizations.}
    \label{tab:1}
\vspace{-2mm}
\end{table*}

Inspired by alignment ~\cite{hong2025robustness} and dispersion ~\cite{chen2025mallowspo}, we decompose the reward function into two corresponding components.
\subsubsection{Alignment Rewards}
In conventional GRPO, accuracy-based rewards directly encourage the model to generate correct outputs by leveraging ground-truth labels. However, it is infeasible to define accuracy rewards in the TTA setting, where supervision is unavailable. To address this issue, we adopt CLIPScore ~\cite{hessel2021clipscore} as the alignment reward that provides label-free signals.

For example, given a test image $x$, a group of outputs that stand for the $C$ class-specific textual prompts will be sampled. To be consistent with Section 3.2, we only consider the top-$K$ classes $\{c_1,\cdots,c_K\}$ and calculate their alignment reward, as:
\begin{equation}
	r_{\mathrm{align}_i} = w \times \max(\mathrm{CLIP}(t_i,v), 0), i\in \{c_1,\cdots,c_K\},
\label{r_1}
\end{equation}
with
\begin{equation}
\mathrm{CLIP}(t,v) = \cos (H_t,H_v),
\end{equation}
where $w=2.5$ is a constant. This reward system motivates the adapted model to enhance the similarity between image representations and semantically relevant textual descriptions.

\subsubsection{Dispersion Rewards}
In test-time adaptation, the absence of supervision makes the model prone to degenerate behaviors, such as assigning nearly identical similarity scores to all candidate outputs or becoming over-confident in noisy predictions. While alignment rewards encourage consistency between visual and textual embeddings, they do not explicitly regulate the relative structure among candidate responses.
To alleviate this issue, we introduce the dispersion reward, which promotes diversity and discriminability among responses. Instead of enforcing uniform probabilities, the dispersion reward explicitly discourages collapsed similarity patterns.

Given an input image $x$, we first apply data augmentation to obtain multiple views of it. Following the protocol of TPT ~\cite{shu2022test}, we select $B$ reliable augmented views based on low self-contropy filtering to ensure stable predictions. For each input image, we sample $K$ textual outputs and compute the corresponding CLIP similarities. Let
\begin{equation}
	\mathrm{sim} = \mathrm{CLIP}(x,t) \in \mathbb{R}^{B \times K},
\end{equation}
where $\mathrm{sim}_{i,j}$ denotes the similarity score between the $i$-th augmented view and the $j$-th sampled textual output. To encourage a moderately dispersed output distribution, we define the dispersion reward of each view as:
\begin{equation}
r_{\mathrm{disp}_i}=\left|\operatorname{sim}_{j, i}-\mu_j\right|, \quad \mu_j=\frac{1}{K} \sum_{i=1}^K \operatorname{sim}_{j, i},
\label{r_2}
\end{equation}

After both components are determined, we compute the final rewards as:
\begin{equation}
	r_i = r_{\mathrm{align}_i} + \lambda r_{\mathrm{disp}_i},
\label{r_3}
\end{equation}
where $\lambda$ is a trade-off hyper-parameter that controls the dispersion of the output distribution.

\subsection{GRPO Loss at Test-Time}
In standard GRPO, a KL-divergence regularization term is introduced to constrain the updated policy to remain close to a reference model. However, \textbf{selecting a strong pretrained reference model may introduce an unintended shortcut during TTA}. Intuitively, the policy update can be dominated by the reference constraint rather than the adaptation signal derived from the test sample itself. To avoid this issue, the objective of our model is computed through Eq. \eqref{eq:grpo} with hyper-parameter $\beta$ fixed to 0. The final optimization loss of GRPO-TTA is formulated as:
\begin{equation}
\resizebox{0.97\linewidth}{!}{$\begin{gathered}
\displaystyle
\mathcal{L}_{\mathrm{GRPO}-\mathrm{TTA}}=-\sum_{i=1}^G \frac{1}{G} \\
\left[\min \left(\frac{\pi_\theta\left(o_i \mid q\right)}{\pi_{\theta_{\text {old }}}\left(o_i \mid q\right)} A_i, \operatorname{cap}\left(\frac{\pi_\theta\left(o_i \mid q\right)}{\pi_{\theta_{\text {old }}}\left(o_i \mid q\right)}, 1-\varepsilon, 1+\varepsilon\right) A_i\right)\right].
\end{gathered}$}
\label{loss}
\end{equation}

To sum up, GRPO-TTA employs the class-specific textual prompts of the dataset as the output group and the redesigned reward function to compute advantages.
The optimization process of our model is summarized in Algorithm~\ref{alg:algorithm}. For simplicity, Algorithm~\ref{alg:algorithm} does not explicitly include the pipeline components that are the same as in TPT.
\section{Experiments}

\begin{table*}[]
\centering
\renewcommand{\arraystretch}{1.0}
\begin{tabular}{lccccccc}
\toprule
Method     & ImageNet       & ImageNet-A     & ImageNet-V2    & ImageNet-R     & ImageNet-Sketch & Average        & OOD Average    \\ \midrule
CLIP$^{\dagger}$ & 66.73          & 47.80          & 60.86          & 73.98          & 46.09           & 59.11          & 57.20          \\
CoOp$^{\dagger}$      & 71.51          & 49.71          & 64.20          & 75.21          & 47.99           & 61.72          & 59.28          \\
CoCoOp$^{\dagger}$    & 71.02          & 50.63          & 64.07          & 76.18          & 48.75           & 62.13          & 59.91          \\ \midrule
TPT$^{\dagger}$       & 68.98          & 54.77          & 63.45          & 77.06          & 47.94           & 62.44          & 60.81          \\
DiffTPT$^{\dagger}$   & 70.30          & 55.68          & 65.10          & 75.00          & 46.80           & 62.28          & 60.65          \\
RLCF$^{\ddagger}$      & \ul{73.30}          & \ul{71.23}          & \ul{67.60}          & \ul{85.34}          & \ul{54.29}           & \ul{70.35}          & \ul{69.62}          \\
WATT$^{\dagger}$      & 64.10          & 50.98          & 62.36          & 75.44          & 48.99           & 60.37          & 59.44          \\
CLIPTTA$^{\dagger}$   & 69.60          & 54.00          & 62.70          & 80.20          & 50.80           & 63.40          & 61.93          \\ \midrule
TDA$^{\dagger}$       & 69.51          & 60.11          & 64.67          & 80.24          & 50.54           & 65.01          & 63.89          \\
DPE$^{\dagger}$       & 71.91          & 59.63          & 65.44          & 80.40          & 52.26           & 65.93          & 64.43          \\
DOTA$^{\dagger}$      & 70.69          & 61.50          & 66.31          & 81.21          & 51.84           & 66.31          & 65.22          \\ \midrule
\rowcolor{gray!15}
GRPO-TTA      & \textbf{74.91} & \textbf{72.97} & \textbf{69.01} & \textbf{85.77} & \textbf{55.17}  & \textbf{71.56} & \textbf{70.73} \\ \bottomrule
\end{tabular}
\caption{Comparison of GRPO-TTA in natural distribution shifts using ViT-B/16 as the backbone. Best results are highlighted in \textbf{bold}, and second-best results are \ul{underlined}. Results annotated with $^{\dagger}$ correspond to results reported in the original papers, whereas $^{\ddagger}$ denote results obtained using our optimizations. The two evaluation metrics, Average and OOD Average, are obtained by computing the mean accuracy across all five datasets and across the four OOD datasets, excluding ImageNet.}
    \label{tab:2}
\vspace{-2mm}
\end{table*}

\begin{table}[]
\centering
\renewcommand{\arraystretch}{1.0}
\begin{tabular}{l||c c c}
\toprule
Method    & Testing Time       & Accuracy   & Gain  \\ \hline
CLIP     & 11 min     & 66.73 & --    \\
TPT$^{\ast}$      & $>$10 h    & 68.98 & $\uparrow$2.25 \\
DiffTPT$^{\ast}$  & $>$20 h    & 70.30 & $\uparrow$3.57 \\
RLCF$^{\ast}$     & $>$10 h    & 73.23 & $\uparrow$6.50 \\
RLCF$^{\star}$     & 3 h 40 min & 73.30 & $\uparrow$6.57 \\
DPE      & 2 h 37 min & 71.91 & $\uparrow$5.18 \\
DOTA     & 32 min     & 70.69 & $\uparrow$3.96 \\
\rowcolor{gray!15}
\textbf{GRPO-TTA$^{\star}$} & 3 h 40 min & \textbf{74.91} & $\uparrow$\textbf{8.18} \\ \bottomrule
\end{tabular}
\caption{Efficiency and accuracy comparison on ImageNet. Note that $^{\ast}$ and $^{\star}$ refer to prompt tuning results and visual tuning results, respectively.}
    \label{tab:3}
\vspace{-2mm}
\end{table}

\subsection{Experimental Settings}
\subsubsection{Datasets}
Following prior works ~\cite{shu2022test,feng2023diverse,karmanov2024efficient,han2024dota}, we conduct our experiments in two scenarios: the cross-domain generalization scenario and the natural distribution shift scenario. For cross-domain generation tasks, we evaluate the performance of the model across 10 diverse image recognition datasets, including FGVCAircraft ~\cite{maji2013fine}, Caltech101 \cite{fei2004learning}, StanfordCars ~\cite{krause20133d}, DTD ~\cite{cimpoi2014describing}, EuroSAT ~\cite{helber2019eurosat}, Flowers102 ~\cite{nilsback2008automated}, Food101 ~\cite{bossard2014food}, OxfordPets ~\cite{parkhi2012cats}, SUN397 ~\cite{xiao2010sun}, and UCF101 ~\cite{soomro2012ucf101}. For the evaluation of the robustness of natural distribution shifts, we utilize ImageNet ~\cite{deng2009imagenet} and four out-of-distribution variants, including ImageNet-A ~\cite{hendrycks2021natural}, ImageNet-V2 ~\cite{recht2019imagenet}, ImageNet-R ~\cite{hendrycks2021many}, and ImageNet-Sketch ~\cite{wang2019learning}.

\subsubsection{Baselines}
We compare the proposed method with the following test-time adaptation approaches: (1) Zero-shot CLIP utilizes an ensemble of 80 prompts ~\cite{radford2021learning}, few shot method CoOp ~\cite{zhou2022learning}, and CoCoOp ~\cite{zhou2022conditional}. (2) Gradient-based training methods, including TPT ~\cite{shu2022test}, DiffTPT ~\cite{feng2023diverse}, RLCF ~\cite{zhao2024test}, WATT ~\cite{osowiechi2024watt}, and CLIPTTA ~\cite{lafon2025cliptta}. (3) Efficient test-time adaptation methods, including TDA ~\cite{karmanov2024efficient}, DPE ~\cite{zhang2024dual}, and DOTA ~\cite{han2024dota}. To ensure a fair comparison, both our method and all baseline models guarantee that the TTA step is set to 1. The results of the above methods are primarily based on the original paper or on rerunning the methods with optimised parameters.

\begin{table}[]
\centering
\renewcommand{\arraystretch}{1.15}
\begin{adjustbox}{width=0.47\textwidth}
\begin{tabular}{lcccc}
\toprule
Method         & ImageNet & ImageNet-A & 10 Cross-domain & Average \\ \midrule
CLIP       & 66.73    & 47.80      & 64.59           & 59.71   \\
RLCF       & 73.30    & 71.23      & 66.01           & 70.18   \\ \midrule
\rowcolor{gray!15}
GRPO-TTA       & \textbf{74.91}    & \textbf{72.97}      & \textbf{70.45}           & \textbf{72.78}   \\
\rowcolor{gray!15}
w/o $r_{disp}$ & 73.31    & 71.63      & 69.71           & 71.55  \\ \bottomrule
\end{tabular}
\end{adjustbox}
\caption{Ablation study to compare the performance of GRPO-TTA with its variant.}
 \label{tab:4}
\vspace{-2mm}
\end{table}

\begin{figure*}[htbp]
\centering
\includegraphics[scale=0.36]{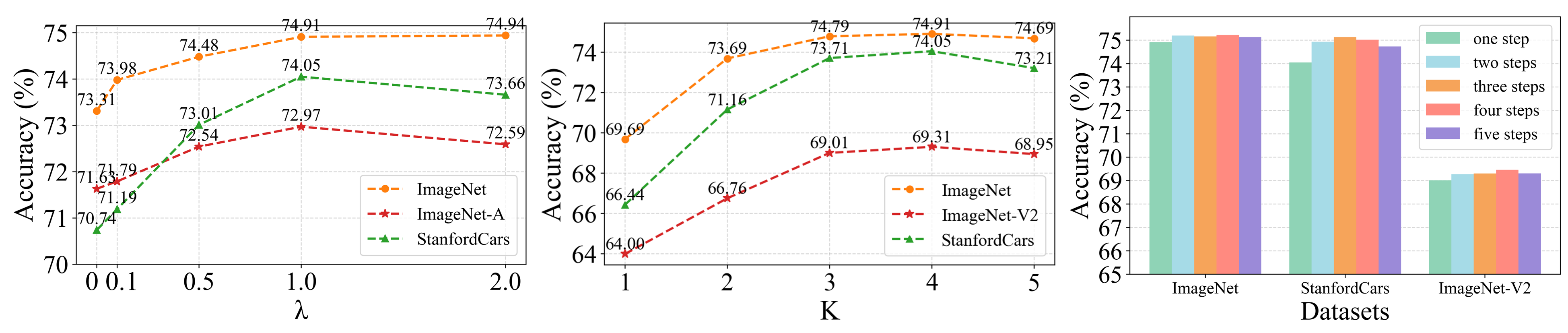}
\caption{Ablation studies. ({\it Left}) Performance on three datasets with varying scale factor $\lambda$ in Eq. \eqref{r_3}; ({\it Middle})  Performance on three datasets with different sampling factors $K$; ({\it Right}) Performance on three datasets with different TTA steps.}
\label{figure2}
\vspace{-2mm}
\end{figure*}

\subsubsection{Implementation Details}
For all experiments, we use ViT-B/16 backbones as the visual encoder of CLIP, with a batch size of 1 to enable online processing. Following previous works ~\cite{shu2022test,zhao2024test,zhang2024dual}, we employ the prompt ensembling strategy. For image augmentation, 63 randomly resized crops are generated for each test sample, consistent with the procedure adopted in DPE \cite{zhang2024dual}. Experiments are conducted in an episodic manner, i.e., restoring the model's parameters after testing each image. By default, the hyper-parameter $\varepsilon$ is set to 0.2, the learning rate is 5e-6, the weight decay value is 5e-4, and the optimizer is AdamW. For most experiments, our model maximizes the reward of the top-4 (sampling factor $K=4$) predictions of each sample with hyper-parameter $\lambda$ set to 1. All experiments are executed on a single NVIDIA RTX 3090 GPU, using top-1 accuracy ($\%$) to measure classification performance.

\subsection{Results and Discussions}
\noindent{\bf{Results on the Cross-Domain Generalization.}} We report the quantitative results of various methods for cross-domain generalization on 10 benchmarks as shown in Table~\ref {tab:1}. Overall, GRPO-TTA achieves the highest average accuracy of 70.45$\%$ and demonstrates superior performance in 5 out of the 10 datasets. Although recent state-of-the-art methods, such as CLIPTTA, DPE, and DOTA, improve performance by updating visual or textual components, they still exhibit performance fluctuations across datasets. For example, CLIPTTA achieves strong results on EuroSAT but performs less favorably on Cars, while DPE excels on Aircraft but shows limited improvement on other benchmarks. In contrast, GRPO-TTA consistently maintains competitive or superior performance across all datasets, achieving improvements of +0.65$\%$, +1.05$\%$, and +0.74$\%$ over CLIPTTA, DPE, and DOTA, respectively. In particular, RLCF is the first attempt to introduce reinforcement learning with human feedback (RLHF) into the test-time adaptation paradigm. GRPO-TTA consistently outperforms RLCF across all 10 datasets, achieving a significant improvement of 4.44$\%$. This indicates that generalizing GRPO to the test-time adaptation setting is both effective and well-motivated.

\noindent{\bf{Results on the Natural Distribution Shifts.}} We further evaluate the performance of GRPO-TTA under natural distribution shifts on ImageNet and its four variants. As shown in Table~\ref{tab:2}, GRPO-TTA consistently achieves the best performance across all datasets, obtaining the highest average accuracy and OOD average accuracy among all compared methods. RLHF-based methods demonstrate clear performance advantages over other methods in the presence of natural distribution shifts.

\noindent{\bf{Efficiency Comparison.}} To illustrate the efficiency of the proposed method, we conduct experiments on the inference time using the ViT-B/16 backbone on the ImageNet dataset. The experimental results are shown in Table~\ref{tab:3}. Compared to the zero-shot CLIP, which achieves 66.73$\%$ top-1 accuracy in 11 minutes, GRPO-TTA yields a notable improvement of +8.18$\%$ after 3 hours and 40 minutes of adaptation. Although TPT and DiffTPT achieve comparable accuracies of 68.98$\%$ and 70.30$\%$, these methods require more than 10 and 20 hours of computation, respectively. Such extended runtimes impose a significant computational overhead, which limits their practical utility in scenarios where efficiency is essential. The efficient TTA method DOTA completes the adaptation process in 32 minutes, but with a corresponding reduced accuracy of 70.69$\%$. The results indicate that GRPO-TTA exhibits optimal performance with moderate adaptation time.

\subsection{Ablation Study}
\noindent{\bf{Effects of Different Rewards.}}
To verify the effectiveness of the proposed method, we investigate the contribution of different reward components. Since the dispersion reward is designed to regulate the distribution induced by the alignment signal, we consider a variant that employs only the alignment reward for comparison. The experimental results are shown in Table~\ref{tab:4}. The experimental results demonstrate that incorporating both alignment and dispersion rewards consistently outperforms the variant that relies solely on alignment rewards. The combination of both provides complementary benefits and contributes to more effective TTA.

\noindent{\bf{Scaling the Dispersion Rewards.}} We also ablate the effect of the dispersion reward by varying the scale factor $\lambda$ in Figure \ref{figure2} ({\it Left}). From the picture, we can see that there is a trade-off between alignment reward and dispersion reward: setting $\lambda$ too high leads to a performance drop. Our experimental results indicate that setting $\lambda = 1.0$ achieves the best performance. 

\noindent{\bf{Effects of Different Sampling Factor.}} When extending GROP to the TTA setting, we need sampling $K$ candidates from the output distribution for reward calculations. As shown in Figure \ref{figure2} ({\it Middle}), experimental performance improves with increasing $K$, but performance starts to decline when $K$ exceeds a certain threshold, e.g., $K=4$. This may be attributed to the reason that too many sampled classes make the optimization process difficult for policy gradient.

\noindent{\bf{Impact of Varing Update Steps.}} To evaluate the effect of TTA steps on performance, we conduct ablation experiments on different numbers of update steps from 1 to 5. As shown in Figure \ref{figure2} ({\it Right}), the number of update steps does not significantly influence performance. Specifically, accuracy improves slightly as the number of TTA steps increases, but begins to degrade beyond a certain point, suggesting that excessive updates may induce overfitting during TTA.

\section{Conclusion}

In this paper, we propose GRPO-TTA, a reinforcement learning framework that adapts Group Relative Policy Optimization to test-time adaptation of vision–language models. By reformulating class-specific prompt prediction as a group-wise policy optimization problem, GRPO-TTA enables effective test-time visual tuning without access to ground-truth labels. We design alignment and dispersion rewards tailored to the TTA setting, which jointly encourage confident semantic alignment while preventing collapsed predictions. Extensive experiments across diverse benchmarks demonstrate that GRPO-TTA consistently outperforms existing test-time adaptation methods under both cross-domain and natural distribution shifts. Despite a moderate increase in computational cost, GRPO-TTA achieves superior accuracy and robust generalization, making it a practical and effective solution for test-time adaptation in VLMs.

\section*{Ethical Statement}

There are no ethical issues.

\bibliographystyle{named}
\bibliography{ijcai26}

\end{document}